\documentclass{bmvc2k}

\title{Dynamic Alignment and Calibration for Multimodal Learning}

\addauthor{Jinghao Xu}{nolle98xu@gmail.com}{1}
\addauthor{Zhenhua Guo}{cszguo@gmail.com}{2}
\addauthor{Xiaofeng Zhu}{seanzhuxf@gmail.com}{3}
\addauthor{Xiaoshuang Shi}
{xsshi2013@gmail.com}{1}

\addinstitution{
School of Computer Science and Engineering, University of Electronic Science and Technology of China,\\
Chengdu,China
}
\addinstitution{
Tianyijiaotong Technology Ltd.,China
}
\addinstitution{
School of Computer Science and Technology, Hainan University,\\
Haikou, China
}

\runninghead{Xu et al.}{Dynamic Alignment and Calibration for Multimodal Learning}

\begin{document}

\maketitle

\begin{abstract}
Dynamic multimodal learning aims to learn robust representations by adaptively modeling information discrepancies across modalities. However, existing methods still suffer from two limitations: (i) static cross-modal alignment strategies usually impose uniform constraints on all samples while overlooking sample-wise variations, potentially leading to unreasonable over-alignment; and (ii) confidence- or uncertainty-aware fusion methods often fail to adequately account for feature magnitude and confidence differences across modalities. For modality pairs with significant feature magnitude differences or small confidence gaps, it might be unreliable to strictly align fusion weights according to confidence. To address these issues, we propose an Alignment- and Calibration-driven Multimodal Learning framework (ACML). Specifically, ACML incorporates a dynamic cross-modal triplet alignment module, which enforces strong semantic consistency for high-confidence positive pairs while encouraging diverse representation learning between high- and low-confidence positive pairs according to their confidence gaps. Additionally, ACML introduces a difference-aware attention calibration strategy that adaptively adjusts attention regularization based on feature magnitude and confidence differences across modalities, thereby mitigating biases caused by unreasonable fusion constraints. Extensive experiments on multiple multimodal benchmark datasets demonstrate that ACML consistently achieves superior performance and robustness over recent state-of-the-art methods.
\end{abstract}

%-------------------------------------------------------------------------
\section{Introduction}
\label{sec:intro}
Multimodal learning has achieved significant success in various fields, such as medical diagnosis\cite{xue2024ai,su2025interpretable, han2025incomplete,zhu2025causal}, emotion recognition\cite{liu2021comparing,geetha2024multimodal,li2025deemo}, and scene understanding\cite{zhou2022mtanet,kong2025multi,wang2025object,wang2024geometric}, by integrating complementary information from diverse modalities to enhance representation power and generalization performance. As the pivotal mechanisms for achieving these advantages, alignment and fusion constitute two core sub-problems in multimodal learning. However, in real-world open environments, data quality and reliability across different modalities often fluctuate dynamically due to device heterogeneity, environmental noise, and occlusions\cite{zhang2024multimodal}. This necessitates multimodal modeling that not only captures cross-modal consistency but also adapts to sample-level variations. Unfortunately, existing methods fall short of adequately modeling such dynamics in both the alignment and fusion stages.

During the alignment stage, most methods still rely on uniform static constraints, applying the same alignment intensity to all samples\cite{li2026multimodal}. This paradigm overlooks the significant disparities in discriminability across different samples\cite{wang2023multi}: while strong alignment helps enhance cross-modal consistency for high-confidence samples that are easy to classify, applying the same intensity to samples with semantic ambiguity or high noise levels may be counterproductive. Forcibly compressing the distance between all positive sample pairs inherently diminishes inter-modal diversity, thereby suppressing the expression of complementary information\cite{qu2024lds2ae,han2025incomplete}. Ultimately, this leads to feature space degradation and undermines the representational advantages that multimodal models are intended to possess.

During the fusion stage, recent studies\cite{han2022trusted,zhang2023provable,cao2024predictive} introduce confidence- or uncertainty-aware dynamic weighting mechanisms to adaptively calibrate modality contributions. However, these methods often overlook feature magnitude differences and confidence differences across modalities. When the feature magnitude difference between modalities is large, their actual contributions to the fused representation are often already significantly differentiated. In such cases, imposing additional strong constraints may lead to unreasonable contribution reallocation and consequently interfere with effective multimodal fusion. On the other hand, when the confidence difference between modalities is small, excessively enforcing confidence-based modulation of modality contributions may restrict potential synergistic modeling across modalities.

\begin{figure}[tbp]
\centering
\includegraphics[width=0.95\textwidth]{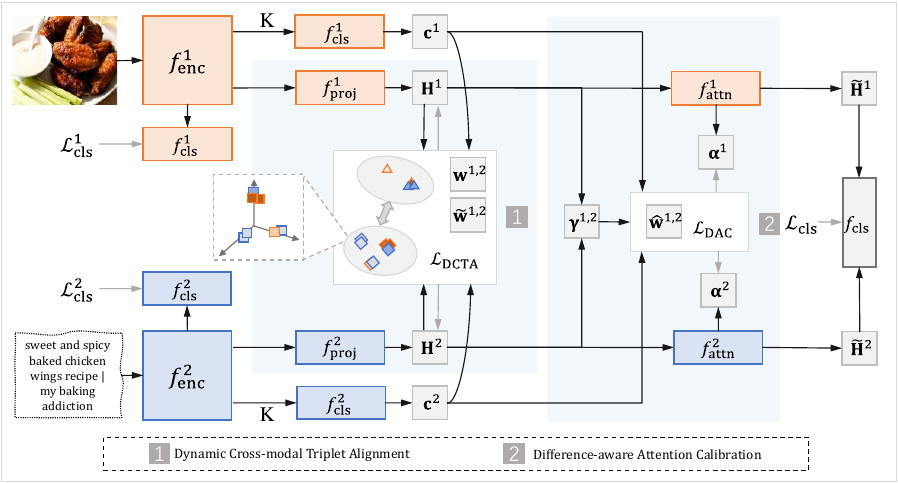}
\caption{Overview of the proposed framework ACML. In $\mathcal{L}_{\text{DCTA}}$, $\triangle$ and $\square$ denote different categories, colors represent modalities, and color intensity reflects confidence.}
\label{fig:overall}
\end{figure}

To address the aforementioned issues, we propose an Alignment- and Calibration-driven Multimodal Learning framework (ACML), as illustrated in Figure~\ref{fig:overall}. ACML consists of two core components: a dynamic cross-modal triplet alignment (DCTA) module and a difference-aware attention calibration (DAC) module. Specifically, modality-specific features are first extracted using dedicated encoders, followed by an independent classification head with multiple stochastic forward passes for sample-level confidence estimation. The features are then projected through independent layers for dimensional alignment and jointly optimized under the proposed dynamic triplet alignment mechanism. This mechanism enhances inter-class discriminability while enforcing cross-modal semantic consistency for high-confidence positive pairs. For low-confidence samples, the alignment constraint is relaxed and an orthogonality encouragement strategy is introduced to preserve potential complementary information and mitigate noise propagation caused by forced alignment. Furthermore, in the DAC module, we impose consistency regularization on the weighted attention distributions across modalities for the same sample, while adaptively adjusting the regularization strength according to feature magnitude differences and confidence differences among modalities. This design enables more reasonable sample-aware fusion while avoiding unnecessary over-regularization on high-magnitude modalities. Extensive experimental results demonstrate that ACML achieves superior effectiveness and robustness under both standard settings and challenging scenarios involving noisy and missing modalities.
\begin{enumerate}
\item We propose ACML, an alignment- and calibration-driven multimodal learning framework that achieves robust and adaptive multimodal representation learning through dynamic cross-modal alignment and difference-aware attention calibration.

\item We design DCTA to enhance semantic consistency for high-confidence positive pairs while promoting diversity between high- and low-confidence pairs, thereby preserving complementary information and mitigating noise propagation.

\item We propose DAC, which dynamically adjusts the strength of attention regularization according to feature magnitude differences and confidence differences, thereby avoiding unreasonable over-regularization on attention weights.
\end{enumerate}

\section{Related Work}
\label{sec:related_work}
\subsection{Multimodal Alignment}
Multimodal alignment aims to establish semantic correspondences across different modalities (e.g., text and images) by mapping their representations into a unified shared embedding space. According to the underlying alignment mechanisms, existing approaches can be broadly categorized into explicit and implicit alignment strategies\cite{li2026multimodal}. Explicit methods \cite{yuan2021multimodal,xu2022multi,liang2022mind} typically rely on similarity matrices to directly measure cross-modal associations, whereas implicit methods\cite{li2022blip,li2023blip,ma2023att} learn joint representations in a latent space through translation or prediction-based objectives. Along the methodological evolution, contrastive learning has emerged as a dominant paradigm for multimodal alignment\cite{liang2022mind}. Its core principle is to pull positive sample pairs closer while pushing negative pairs apart, thereby enforcing discriminative cross-modal consistency in the embedding space \cite{yuan2021multimodal,dufumier2024align}. Our method follows the explicit alignment paradigm but addresses the limited consideration of modality dynamics in prior work by introducing a dynamic-aware mechanism. Moreover, we extend conventional contrastive objectives by designing a triplet alignment loss, which aims to strike a better balance between promoting cross-modal consistency and preserving complementary information.
\subsection{Multimodal Fusion}
Multimodal fusion aims to integrate complementary information from different modalities to improve performance on downstream tasks. Existing approaches are typically categorized into early, intermediate, or late fusion based on the fusion stage \cite{atrey2010multimodal}. Early fusion methods \cite{zeng2024disentanglement,suo2025automatic} combine multiple modalities directly at the data level, but they often struggle with high-dimensional or heterogeneous data. Intermediate fusion approaches \cite{ding2024multimodal} focus on learning joint feature representations across modalities, while late fusion methods \cite{zhang2023provable,cao2024predictive} emphasize semantic-level integration of individual modality predictions.

Recent studies have increasingly focused on dynamic multimodal fusion \cite{nagrani2021attention,xue2023dynamic,zheng2023multi,zhang2024multimodal} to address the observation that the relative importance of each modality may vary across samples or over time. Unlike static fusion strategies that assume fixed modality weights regardless of input conditions, dynamic fusion approaches leverage attention mechanisms or uncertainty modeling to adaptively modulate the contribution of each modality. This adaptability enables them to better handle challenging scenarios, such as degraded data quality or cross-domain variations, thereby improving both robustness and generalization. However, existing studies typically impose homogeneous constraints on fusion weights while overlooking the differences in feature norms and confidence across modalities. This oversight may lead to unreasonable constraints on modality pairs with abnormal norm ratios or highly consistent confidence, thereby suppressing the expression of complementary cross-modal information. In contrast, we introduce a pairwise constraint loss that adaptively adjusts the constraint strength for different modality pairs based on both norm and confidence differences.
\section{Method}
\subsection{Feature Extraction and Confidence Estimation}
Given a mini-batch of \( N \) multimodal samples, each of which has \( M \) modality-specific inputs and a corresponding class label. The inputs are represented as \( \{\mathbf{X}^m\}_{m=1}^M \), where \( \mathbf{X}^m = \{\mathbf{x}_i^m\}_{i=1}^N \) denotes the \( m\)-th modality of $N$ samples. The associated label set is \( \mathbf{Y} = \{y_1, y_2, \dots, y_N\} \), with \( y_i \in \{1, \dots, C\} \).

For the \(m\)-th modality, the input \( \mathbf{X}^m \) is first passed through a modality-specific encoder \( f^m_\text{enc}(\cdot) \) to extract high-level feature representations:
\begin{equation}
    \mathbf{Z}^m = f^m_\text{enc}\left(\mathbf{X}^m\right),
\end{equation}
where  $\mathbf{Z}^m = [\mathbf{z}_1^m, \mathbf{z}_2^m, \dots, \mathbf{z}_N^m]^\top \in \mathbb{R}^{N \times d_m}$ is the feature matrix, $\mathbf{z}_i^m \in \mathbb{R}^{d_m}$ represents the feature vector of the $i$-th sample. 

To obtain robust and stable confidence estimates during training, we employ a stochastic inference scheme. Specifically, we leverage the classifier $f_{\text{cls}}^m(\cdot)$ to construct two parallel computational paths. In the primary training path, the feature $\mathbf{z}_i^m$ is directly fed into the classifier without activating dropout. This path is supervised by the standard cross-entropy loss $\mathcal{L}_{\text{cls}}^m$, which is to ensure that the model learns fundamental discriminative representations:
\begin{equation}
\mathcal{L}_{\text{cls}}^m = - \frac{1}{N} \sum_{i=1}^N \log \left( \left[ s(f_{\text{cls}}^m(\mathbf{z}_i^m)) \right]_{y_i} \right),
\end{equation}
where $s(\cdot)$ denotes the softmax activation function, and $[\cdot]_{y_i}$ extracts the probability corresponding to the ground-truth index. 
Meanwhile, to mitigate the prevalent over-confidence issue associated with a single deterministic forward pass and to enhance the quality of confidence estimation, we introduce a probabilistic auxiliary path. Under this path, we apply a dropout operator $\delta(\cdot)$ to the same feature vector $\mathbf{z}_i^m$ and execute $K$ stochastic forward passes. In this work, we set the dropout probability to $0.1$ and the number of stochastic forward passes $K$ to $10$. For each pass $k \in \{1, \dots, K\}$, the generated stochastic prediction vector is formulated as:
\begin{equation}
\mathbf{{p}}_{i,k}^m = s(f_{\text{cls}}^m(\delta(\mathbf{z}_i^m))).
\end{equation}
The final confidence score $c_i^m$ for the $i$-th sample is defined as the empirical mean of the ground-truth probabilities across these $K$ stochastic predictions:
\begin{equation}
c_i^m = \frac{1}{K} \sum_{k=1}^K \left[ \mathbf{{p}}_{i,k}^m \right]_{y_i}.
\end{equation}
After obtaining the confidence vector $\mathbf{c}^m = [c_1^m, c_2^m, \dots, c_N^m]^\top \in \mathbb{R}^N$, we detach it from the computational graph to serve as dynamic guidance for subsequent modules.

\subsection{Dynamic Cross-modal Triplet Alignment}
Cross-modal semantic alignment is a fundamental component of multimodal learning \cite{li2026multimodal}. However, due to the varying discriminative capabilities across modalities and the need to preserve complementary information, enforcing uniform semantic consistency constraints on all sample pairs is often suboptimal. To address this issue, we propose a Dynamic Cross-modal Alignment loss, denoted as $\mathcal{L}_{\mathrm{DCTA}}$. The proposed loss dynamically adjusts alignment strength according to sample confidence and explicitly preserves feature separability based on confidence discrepancy, thereby alleviating noisy semantic aggregation and promoting feature diversity.

For the feature representation $\mathbf{Z}^m$ of the $m$-th modality, we first employ a modality-specific projection layer $f_{\text{proj}}^m(\cdot)$, implemented as an MLP, to map the features into a shared multimodal embedding space:
\begin{equation}
\mathbf{H}^m
=
f_{\text{proj}}^m(\mathbf{Z}^m)
=
\{\mathbf{h}_1^m, \mathbf{h}_2^m, \dots, \mathbf{h}_N^m\},
\end{equation}
where $\mathbf{h}_i^m \in \mathbb{R}^d$ denotes the embedded representation of the $i$-th sample in the $m$-th modality.

Within the embedding space, we compute the cosine similarity matrix $\mathbf{S}^{m,n} \in \mathbb{R}^{N \times N}$ for each modality pair $(m,n)$, where each element $s_{ij}^{m,n}$ is defined as:
\begin{equation}
s_{ij}^{m,n}
=
\frac{
\mathbf{h}_i^{m\top}\mathbf{h}_j^n
}{
\|\mathbf{h}_i^m\|
\cdot
\|\mathbf{h}_j^n\|
}.
\end{equation}

To dynamically regulate the alignment strength according to paired-sample reliability, we define a joint confidence weight:
\begin{equation}
w_{ij}^{m,n}
=
\min(c_i^m, c_j^n).
\end{equation}
In this way, high-confidence positive pairs are encouraged to achieve stronger semantic consistency, while unreliable pairs are softly relaxed to alleviate noisy semantic aggregation. The alignment term is formulated as:
\begin{equation}
\mathrm{Align}_{i}^{m,n}
=
\sum_{j=1}^{N}
\mathbb{I}(y_i = y_j)
\,
w_{ij}^{m,n}
\,
\exp\left(
\frac{s_{ij}^{m,n}}{\tau}
\right),
\end{equation}
where $\mathbb{I}(\cdot)$ denotes the indicator function and $\tau$ is a temperature parameter.

Although the confidence-weighted alignment effectively alleviates noisy aggregation caused by unreliable samples, it does not explicitly preserve the intrinsic representation diversity within the same class. To this end, we further introduce a confidence-gap-aware decorrelation constraint to encourage more diverse feature representations among positive pairs. Specifically, we define a confidence-gap weight as:
\begin{equation}
\tilde{w}_{ij}^{m,n}
=
|c_i^m - c_j^n|.
\end{equation}
This weight adaptively modulates the strength of the decorrelation constraint according to the confidence discrepancy between positive pairs. Based on this weighting scheme, we further construct an orthogonality-inspired decorrelation objective using the absolute feature similarity. Positive pairs with larger confidence gaps are subjected to stronger decorrelation, encouraging them to maintain lower feature correlation. As a result, the learned representations remain semantically consistent while preserving sufficient diversity, thereby retaining potential complementary information. The decorrelation term is defined as:
\begin{equation}
\mathrm{Decor}_{i}^{m,n}
=
\sum_{j=1}^{N}
\mathbb{I}(y_i = y_j)
\,
\tilde{w}_{ij}^{m,n}
\,
\exp\left(
\frac{|s_{ij}^{m,n}|}{\tau}
\right).
\end{equation}

For negative pairs ($y_i \neq y_j$), we further introduce a repulsion term to separate irrelevant category representations in the embedding space:
\begin{equation}
\mathrm{Repel}_{i}^{m,n}
=
\sum_{j=1}^{N}
\mathbb{I}(y_i \neq y_j)
\,
\exp\left(
\frac{s_{ij}^{m,n}}{\tau}
\right).
\end{equation}

Finally, we integrate the alignment, decorrelation, and repulsion terms into a unified optimization objective. The batch-wise loss between modality $m$ and modality $n$ is:
\begin{equation}
\mathcal{L}_{\mathrm{DCTA}}^{m,n}
=
-
\frac{1}{N}
\sum_{i=1}^{N}
\log
\frac{
\mathrm{Align}_{i}^{m,n}
}{
\mathrm{Align}_{i}^{m,n}
+
\mathrm{Decor}_{i}^{m,n}
+
\mathrm{Repel}_{i}^{m,n}
}.
\end{equation}
The final global objective is obtained by summing all bidirectional modality-pair losses:
\begin{equation}
\mathcal{L}_{\mathrm{DCTA}}
=
\sum_{1 \le m < n \le M}
\left(
\mathcal{L}_{\mathrm{DCTA}}^{m,n}
+
\mathcal{L}_{\mathrm{DCTA}}^{n,m}
\right).
\end{equation}

\subsection{Difference-aware Attention Calibration}
In dynamic multimodal fusion, modality contributions are typically modulated by predictive confidence or uncertainty \cite{han2022multimodal,zhang2023provable,cao2024predictive}. Existing methods typically adopt a unified global attention calibration strategy without adequately considering sample-wise modality differences. However, when the feature magnitude difference between modalities is large, their contributions to the fused representation are often already significantly differentiated, making additional strong calibration potentially unnecessary and even harmful to effective multimodal fusion. Meanwhile, when the confidence difference between modalities is small, excessive calibration may suppress potential synergistic information across modalities. To address these issues, we propose a difference-aware attention calibration constraint, termed $\mathcal{L}_{\mathrm{DAC}}$, which adaptively emphasizes modality pairs with distinguishable confidence levels and comparable feature magnitudes, thereby enabling more adaptive and effective multimodal fusion.

Specifically, for the $i$-th sample, we employ a modality-specific attention layer $f^m_{\text{attn}}$, implemented using a modality-specific MLP followed by a Sigmoid function, to compute the attention score for each modality. The original features are then reweighted by the corresponding attention scores to obtain the attention-weighted features $\tilde{\mathbf{h}}_i^m$, as follows:
\begin{equation}
\alpha_i^m,\tilde{\mathbf{h}}_i^m = f_{\text{attn}}^m(\mathbf{h}_i^m).
\end{equation}

To calibrate the relative attention allocation at the sample level, we define the pairwise attention ratio and confidence ratio for each modality pair $(m,n)$ as:
\begin{equation}
\mathrm{AttnRatio}_{i}^{m,n}
=
\frac{\alpha_i^{m}}
{\alpha_i^{m} + \alpha_i^{n} + \varepsilon},
\quad
\mathrm{ConfRatio}_{i}^{m,n}
=
\frac{c_i^{m}}
{c_i^{m} + c_i^{n} + \varepsilon},
\end{equation}
where $\mathrm{AttnRatio}_{i}^{m,n}$ measures the relative attention allocation between modalities, while $\mathrm{ConfRatio}_{i}^{m,n}$ represents their relative predictive confidence. Here, $\varepsilon$ is a small constant introduced for numerical stability.

Furthermore, to selectively enforce reliable calibration signals, we define a difference-aware pairwise weight as:
\begin{equation}
\hat{w}_i^{m,n}
=
\gamma_i^{m,n}
\cdot
\left(
|c_i^m-c_i^n|+1
\right),
\end{equation}
where
$
\gamma_i^{m,n}
=
\frac{
\min(\|\mathbf h_i^m\|,\|\mathbf h_i^n\|)
}{
\max(\|\mathbf h_i^m\|,\|\mathbf h_i^n\|)+\varepsilon
}$ measures the similarity between modality magnitudes. When the magnitude difference between modalities is large, the corresponding calibration constraint is adaptively relaxed to avoid unreasonable contribution reallocation. In contrast, modality pairs with comparable feature magnitudes receive relatively larger calibration weights. Meanwhile, the confidence difference term $(|c_i^m-c_i^n|+1)$ preserves a basic alignment objective while further emphasizing modality pairs with more distinguishable reliability levels. Therefore, $\hat{w}_i^{m,n}$ jointly models feature magnitude differences and confidence differences, enabling more adaptive and reliable attention calibration.

Finally, the difference-aware attention calibration constraint between modality $m$ and modality $n$ is formulated as:
\begin{equation}
\mathcal{L}_{\mathrm{DAC}}^{m,n}
=
\frac{1}{N}
\sum_{i=1}^{N}
\hat{w}_i^{m,n}
\left(
\mathrm{AttnRatio}_{i}^{m,n}
-
\mathrm{ConfRatio}_{i}^{m,n}
\right)^2.
\end{equation}
The global difference-aware attention calibration constraint is obtained by summing over all modality pairs:
\begin{equation}
\mathcal{L}_{\mathrm{DAC}}
=
\sum_{1 \le m < n \le M}
\mathcal{L}_{\mathrm{DAC}}^{m,n}.
\end{equation}
\subsection{Fusion and Optimization Objective}

After cross-modal alignment and attention calibration, we adopt a mean fusion strategy to aggregate all reweighted modality representations and perform final classification through the classification head $f_{\mathrm{cls}}(\cdot)$:
\begin{equation}
\mathbf{h}_i^{\mathrm{fuse}}
=
\frac{1}{M}
\sum_{m=1}^{M}
\tilde{\mathbf{h}}_i^m,
\quad
\mathcal{L}_{\mathrm{cls}}
=
-
\frac{1}{N}
\sum_{i=1}^{N}
\log
\left(
\left[
s
\left(
f_{\mathrm{cls}}
(
\mathbf{h}_i^{\mathrm{fuse}}
)
\right)
\right]_{y_i}
\right).
\end{equation}

Finally, the overall optimization objective of ACML is formulated as
\begin{equation}
\mathcal{L}
=
\mathcal{L}_{\mathrm{CLS}}
+
\lambda_1
\mathcal{L}_{\mathrm{DCTA}}
+
\lambda_2
\mathcal{L}_{\mathrm{DAC}},
\end{equation}
where $\mathcal{L}_{\mathrm{CLS}} = \mathcal{L}_{\mathrm{cls}} + \sum_{m=1}^{M}\mathcal{L}_{\mathrm{cls}}^m$ denotes the summation of the fused classification loss and all unimodal classification losses, while $\lambda_1$ and $\lambda_2$ are balancing hyperparameters for $\mathcal{L}_{\mathrm{DCTA}}$ and $\mathcal{L}_{\mathrm{DAC}}$, respectively. For clarity, we summarize the whole training procedure of ACML in Algorithm \ref{algo}.
%% 写算法伪代码或者流程的前期准备
\renewcommand{\algorithmicrequire}{\textbf{Input:}}  % Use Input in the format of Algorithm
\renewcommand{\algorithmicensure}{\textbf{Output:}} % Use Output in the format of Algorithm
\renewcommand{\thealgorithm}{1}
\begin{algorithm}[htbp]
  \caption{ACML} % 名称
  \small
  \begin{algorithmic}[1]
    \Require Training data $\mathcal{D}$, the number of training iterations $T$.
    \Ensure Well trained ACML. 
    \For {iteration $\in$ [1, $T$]}
      \State Randomly select a batch of samples $\{\{\mathbf{X}^m\}^M_{m=1},\mathbf{Y}\}$ from $\mathcal{D}$;
      \State \textcolor{gray}{\# Obtain modality-specific representation.}
      \State $\{\mathbf{Z}^m\}^M_{m=1} \leftarrow \{f^m_\text{enc}(\mathbf{X}^m)\}^M_{m=1}$ 
      \State \textcolor{gray}{\# Get the confidence score for each modal sample.}
      \State $\{ \mathbf{c}^{m}\}_{m=1}^M \leftarrow$ Eqs. (2)-(4)
      \State \textcolor{gray}{\# Map multimodal features to the same dimension.}
      \State $\{\mathbf{H}^m\}^M_{m=1} \leftarrow \{f^m_\text{proj}(\mathbf{Z}^m)\}^M_{m=1}$ 
      \State \textcolor{gray}{\# Perform dynamic cross-modal triplet alignment.}
      \State $\mathcal{L}_\text{DCTA} \leftarrow$ Eqs. (6)-(13)
      \State \textcolor{gray}{\# Compute attention weights and reweight features.}
      \State $\alpha^m,\tilde{\mathbf{H}}^m  \leftarrow f_{\text{attn}}^m(\mathbf{H}^m)$
      \State \textcolor{gray}{\# Perform difference-aware attention calibration.}
      \State $\mathcal{L}_\text{DAC} \leftarrow$ Eqs. (15)-(18)
      \State \textcolor{gray}{\# Perform multimodal fusion and final classification.}
      \State ${\mathbf{H}^\text{fuse},\mathcal{L}_\text{cls}} \leftarrow$ Eq. (19)
      \State \textcolor{gray}{\# Integrate multiple constraints into the final optimization objective.}
      \State $\mathcal{L} \leftarrow$ Eq. (20)
    \State  Back-propagate $\mathcal{L}$ to update model parameters;
    \EndFor
  \end{algorithmic}
  \label{algo}
\end{algorithm}
\section{Experiments and Analysis}
In this section, we evaluate the effectiveness of the proposed ACML by addressing the following six Research Questions (RQs):
\vspace{-6pt}
\begin{itemize}
    \setlength{\itemsep}{2pt}
    \setlength{\parskip}{0pt}
    \setlength{\parsep}{0pt}

\item \textbf{RQ1 (Effectiveness \& Robustness)}: How does ACML compare with SOTA baselines in terms of overall performance and robustness against noise?

\item \textbf{RQ2 (Ablation Study)}: What are the individual contributions of the key components within ACML to its overall performance?

\item \textbf{RQ3 (Visualization Analysis)}: How does $\mathcal{L}_{\mathrm{DCTA}}$ influence the cross-modal feature distribution in the embedding space?

\item \textbf{RQ4 (Hyper-parameter Analysis)}: How do the critical hyper-parameters influence the stability of ACML?

\item \textbf{RQ5 (Applicability to Missing Modalities)}: Can ACML effectively enhance prompt-tuning frameworks as a plug-and-play module in challenging missing-modality scenarios?

\item \textbf{RQ6 (Scalability to Multiple Modalities)}: Does ACML maintain its effectiveness and scalability when extended to complex scenarios with more than two modalities ($M > 2$)?

\end{itemize}
\vspace{-6pt}
\subsection{RQ1 (Effectiveness \& Robustness)}
\begin{table}[t!]
\small
\centering
\caption{Classification comparison when 50\% of the modalities are corrupted with Gaussian noise, i.e., zero mean with variance of $\epsilon$. We highlight the optimal results using bold text and indicate the suboptimal results with underlining for clarity.}
\resizebox{\textwidth}{!}{
\begin{tabular}{c|c|cccccc}
\toprule
\multirow{2}{*}{Dataset} & \multirow{2}{*}{Method} 
& \multicolumn{2}{c}{$\epsilon = \mathbf{0.0}$} 
& \multicolumn{2}{c}{$\epsilon = \mathbf{5.0}$} 
& \multicolumn{2}{c}{$\epsilon = \mathbf{10.0}$} \\
& & Avg±Std & Worst & Avg±Std & Worst & Avg±Std & Worst \\
\midrule

\multirow{10}{*}{\makecell{NYU\\Depth V2}} 

& Depth 
& 61.97{\scriptsize ±0.99} & 60.40 
& 49.65{\scriptsize ±3.22} & 43.27 
& 42.42{\scriptsize ±2.42} & 36.85 \\

& RGB 
& 62.57{\scriptsize ±0.82} & 61.32 
& 52.56{\scriptsize ±2.28} & 47.55 
& 46.16{\scriptsize ±3.10} & 38.84 \\

& LateFusion 
& 69.22{\scriptsize ±0.98} & 67.43 
& 56.35{\scriptsize ±2.84} & 51.22 
& 48.65{\scriptsize ±3.40} & 42.36 \\

& Concat 
& 68.70{\scriptsize ±0.67} & 67.89 
& 55.76{\scriptsize ±2.23} & 49.54 
& 48.17{\scriptsize ±3.36} & 40.21 \\

& TMC 
& \underline{71.42{\scriptsize ±0.84}} & \underline{70.03}
& 61.44{\scriptsize ±2.19} & 56.58
& 53.36{\scriptsize ±3.36} & 45.87 \\

& QMF 
& 68.55{\scriptsize ±1.29} & 66.82
& 56.60{\scriptsize ±3.12} & 50.31
& 49.04{\scriptsize ±3.69} & 41.44 \\

& PDF 
& 68.46{\scriptsize ±0.69} & 66.67
& 56.67{\scriptsize ±2.69} & 48.47
& 48.61{\scriptsize ±3.56} & 38.23 \\

& LFM  
& 69.27{\scriptsize ±0.62} & 68.20
& 59.17{\scriptsize ±2.15} & 53.98
& 51.35{\scriptsize ±2.70} & 44.95 \\

& OPM 
& 71.35{\scriptsize ±0.93} & 69.88
& \underline{63.35{\scriptsize ±1.41}} & \underline{60.40}
& \underline{54.82{\scriptsize ±2.37}} & \underline{49.54} \\

& \cellcolor{gray!20}\textbf{ACML}  
& \cellcolor{gray!20}\textbf{72.72{\scriptsize ±0.81}}
& \cellcolor{gray!20}\textbf{71.56}
& \cellcolor{gray!20}\textbf{65.43{\scriptsize ±1.53}}
& \cellcolor{gray!20}\textbf{60.86}
& \cellcolor{gray!20}\textbf{58.11{\scriptsize ±2.13}}
& \cellcolor{gray!20}\textbf{49.69} \\

\midrule

\multirow{10}{*}{\makecell{SUN\\RGB-D}} 

& Depth 
& 52.58{\scriptsize ±0.67} & 51.43
& 39.05{\scriptsize ±1.06} & 37.15
& 34.66{\scriptsize ±1.61} & 31.08 \\

& RGB 
& 57.46{\scriptsize ±0.39} & 56.84
& 48.05{\scriptsize ±0.88} & 45.25
& 42.45{\scriptsize ±1.47} & 38.74 \\

& LateFusion 
& 59.90{\scriptsize ±0.45} & 59.24
& 47.78{\scriptsize ±2.03} & 43.01
& 41.27{\scriptsize ±2.82} & 35.57 \\

& Concat 
& 59.52{\scriptsize ±0.50} & 58.58
& 48.54{\scriptsize ±1.63} & 45.12
& 42.74{\scriptsize ±1.98} & 38.96 \\

& TMC 
& 61.02{\scriptsize ±0.37} & 60.34
& 50.64{\scriptsize ±1.69} & 47.18
& 43.89{\scriptsize ±3.22} & 37.54 \\

& QMF 
& 61.11{\scriptsize ±0.26} & 60.59
& 51.44{\scriptsize ±1.06} & 49.13
& 46.45{\scriptsize ±1.61} & \underline{42.71} \\

& PDF 
& 60.89{\scriptsize ±0.39} & 60.23
& 50.24{\scriptsize ±1.43} & 46.71
& 44.92{\scriptsize ±2.21} & 40.44 \\

& LFM 
& 61.98{\scriptsize ±0.25} & 61.64
& \underline{52.84{\scriptsize ±0.96}} & \underline{50.55}
& \underline{47.36{\scriptsize ±1.72}} & 42.58 \\

& OPM 
& \underline{62.23{\scriptsize ±0.26}} & \underline{61.79}
& 51.72{\scriptsize ±1.08} & 49.60
& 43.62{\scriptsize ±1.90} & 40.59 \\

& \cellcolor{gray!20}\textbf{ACML} 
& \cellcolor{gray!20}\textbf{63.25{\scriptsize ±0.25}}
& \cellcolor{gray!20}\textbf{62.76}
& \cellcolor{gray!20}\textbf{56.50{\scriptsize ±0.87}}
& \cellcolor{gray!20}\textbf{54.15}
& \cellcolor{gray!20}\textbf{49.90{\scriptsize ±1.15}} 
& \cellcolor{gray!20}\textbf{47.48} \\

\midrule

\multirow{10}{*}{\makecell{FOOD\\101}} 

& Img 
& 64.78{\scriptsize ±0.17} & 64.54
& 34.55{\scriptsize ±0.58} & 33.66
& 32.99{\scriptsize ±0.28} & 32.54 \\

& Text 
& 86.49{\scriptsize ±0.05} & 86.44
& 67.35{\scriptsize ±0.18} & 67.04
& 43.84{\scriptsize ±0.31} & 43.39 \\

& LateFusion 
& 90.71{\scriptsize ±0.09} & 90.61
& 70.07{\scriptsize ±2.86} & 64.56
& 59.51{\scriptsize ±0.87} & 57.81 \\

& Concat 
& 88.35{\scriptsize ±0.57} & 87.68
& 63.28{\scriptsize ±2.35} & 59.45
& 51.18{\scriptsize ±3.71} & 46.34 \\

& TMC 
& 90.26{\scriptsize ±0.17} & 89.94
& 73.69{\scriptsize ±0.82} & 72.15
& 61.51{\scriptsize ±0.21} & 61.01 \\

& QMF 
& 92.88{\scriptsize ±0.11} & 92.70
& \underline{76.39{\scriptsize ±0.62}} & 75.30
& 61.86{\scriptsize ±0.41} & 61.74 \\

& PDF 
& \underline{92.98{\scriptsize ±0.17}} & 92.71
& 75.34{\scriptsize ±0.55} & 74.29
& \underline{62.77{\scriptsize ±0.23}} & \underline{62.45} \\

& LFM 
& 92.80{\scriptsize ±0.05} & \underline{92.74}
& 76.15{\scriptsize ±0.35} & \underline{75.51}
& 62.57{\scriptsize ±0.34} & 61.88 \\

& OPM 
& 92.60{\scriptsize ±0.22} & 92.21
& 75.17{\scriptsize ±0.51} & 74.37
& 60.97{\scriptsize ±0.78} & 59.66 \\

& \cellcolor{gray!20}\textbf{ACML} 
& \cellcolor{gray!20}\textbf{93.36{\scriptsize ±0.10}}
& \cellcolor{gray!20}\textbf{93.23}
& \cellcolor{gray!20}\textbf{76.64{\scriptsize ±0.49}}
& \cellcolor{gray!20}\textbf{75.69}
& \cellcolor{gray!20}\textbf{63.15{\scriptsize ±0.24}}
& \cellcolor{gray!20}\textbf{62.76} \\

\bottomrule
\end{tabular}
}
\label{generalization_ability}
\end{table}
\textbf{Setting.} We evaluate ACML on two representative multimodal tasks: (1) \textbf{Scene Recognition} on the NYU Depth V2\cite{silberman2012indoor} and SUN RGB-D \cite{song2015sun} datasets, and (2) \textbf{Image–Text Classification} on the UPMC FOOD101 \cite{wang2015recipe} dataset. Detailed dataset descriptions are provided in the supplementary material. All models are implemented in PyTorch and optimized using Adam on an NVIDIA A100 GPU. For the scene recognition task, we adopt ResNet-18 as the backbone network, with a learning rate of $1\times10^{-4}$, a batch size of 16, and an embedding dimension $d$ of 512. For the image--text classification task, ResNet-152 and BERT are used to extract image and text features, respectively, with a learning rate of $5\times10^{-5}$, an effective batch size of 640 achieved via gradient accumulation (i.e., batch size 32 with 20 accumulation steps), and an embedding dimension $d$ of 2048. Following \cite{ma2021trustworthy}, we simulate real-world multimodal uncertainty by injecting Gaussian noise into images and blank noise into text. Scene recognition experiments are repeated over 10 random seeds, while image–text classification experiments are conducted with 5 random seeds. We report both the average accuracy (with standard deviation) and the worst-case accuracy. All models are trained for up to 100 epochs, with early stopping applied to prevent overfitting.

\textbf{Analysis.} As shown in Table \ref{generalization_ability}, we compare ACML with three categories of methods: traditional fusion approaches (\textbf{Concat} and \textbf{LateFusion}), state-of-the-art dynamic learning methods (\textbf{TMC}\cite{han2022trusted}, \textbf{QMF}\cite{zhang2023provable}, and \textbf{PDF}\cite{cao2024predictive}), and modality-imbalance learning methods (\textbf{LFM}\cite{yang2024facilitating} and \textbf{OPM}\cite{10694738}). The results show that ACML consistently achieves the best performance across all datasets under both clean and noisy conditions. Specifically, under the clean setting ($\epsilon=0$), ACML yields relative gains of 1.82\%, 1.64\%, and 0.41\% in average accuracy over the strongest baseline on NYU Depth V2, SUN RGB-D, and FOOD101, respectively. As the noise level increases, the advantage of ACML becomes more pronounced on NYU Depth V2 and SUN RGB-D. For example, under severe corruption ($\epsilon=10$), the relative gains in average accuracy further increase to 6.00\% and 5.36\%, respectively. This trend suggests that ACML not only improves predictive performance under clean conditions but also exhibits stronger robustness against modality corruption.

\subsection{RQ2 (Ablation Study)}
\begin{table*}[tbp]
\centering
\small
\caption{Ablation studies on NYU Depth V2. We report both the average performance and the worst-case performance under different noise levels.}
\label{tab:ablation_study}

\resizebox{\textwidth}{!}{
\begin{tabular}{l|cc|cc|cc}
\toprule

\multirow{2}{*}{Configuration}

& \multicolumn{2}{c}{$\epsilon = \mathbf{0.0}$}
& \multicolumn{2}{c}{$\epsilon = \mathbf{5.0}$}
& \multicolumn{2}{c}{$\epsilon = \mathbf{10.0}$} \\

\cmidrule(lr){2-3}
\cmidrule(lr){4-5}
\cmidrule(lr){6-7}

& Avg±Std & Worst
& Avg±Std & Worst
& Avg±Std & Worst \\

\midrule

Baseline
& 70.87{\scriptsize±0.79}
& 69.27
& 62.61{\scriptsize±1.53}
& 58.87
& 55.02{\scriptsize±3.38}
& 46.79 \\

Baseline + $\mathcal{L}_{\text{DCTA}}$
& \underline{72.37{\scriptsize±0.80}}
& \underline{70.80}
& \underline{63.82{\scriptsize±1.97}}
& \underline{58.56}
& \underline{55.70{\scriptsize±2.91}}
& \underline{48.62} \\

Baseline + $\mathcal{L}_{\text{DAC}}$
& 71.67{\scriptsize±0.82}
& 70.32
& 63.66{\scriptsize±2.58}
& 58.42
& 53.93{\scriptsize±4.33}
& 47.59 \\

\rowcolor{gray!20}
ACML
& \textbf{72.72{\scriptsize±0.71}}
& \textbf{71.56}
& \textbf{65.43{\scriptsize±1.63}}
& \textbf{60.86}
& \textbf{58.11{\scriptsize±2.69}}
& \textbf{49.69} \\

\midrule

ACML ($\mathcal{L}_{\text{DCTA}}
\rightarrow
\mathcal{L}_{\text{SC}}$)

& 72.48{\scriptsize±0.69}
& 71.10
& 65.06{\scriptsize±1.74}
& 60.86
& 57.09{\scriptsize±2.75}
& 51.07 \\

ACML ($\mathcal{L}_{\text{DCTA}}
\rightarrow
\mathcal{L}_{\text{CSC}}$)

& 72.55{\scriptsize±0.76}
& 71.41
& 64.67{\scriptsize±1.55}
& 60.09
& 57.43{\scriptsize±2.39}
& 51.38 \\

\midrule

ACML ($\mathcal{L}_{\text{DAC}}
\rightarrow
\mathcal{L}_{\text{MSE}}$)
& 72.49{\scriptsize±0.71}
& 71.10
& 64.78{\scriptsize±2.57}
& 58.72
& 57.41{\scriptsize±3.15}
& 49.69 \\

\bottomrule
\end{tabular}
}

\end{table*}
\textbf{Setting.} We conduct ablation studies on the NYU Depth V2 dataset to systematically analyze the contributions of the dynamic cross-modal triplet alignment loss ($\mathcal{L}_{\text{DCTA}}$) and the difference-aware attention constraint ($\mathcal{L}_{\text{DAC}}$). Furthermore, based on the final ACML framework, we further investigate the effectiveness of different optimization objectives and constraint formulations. Specifically, we replace $\mathcal{L}_{\text{DCTA}}$ with the standard supervised contrastive loss ($\mathcal{L}_{\text{SC}}$) and the confidence-constrained supervised contrastive loss ($\mathcal{L}_{\text{CSC}}$), respectively. Meanwhile, we replace the proposed $\mathcal{L}_{\text{DAC}}$ with a standard unweighted MSE constraint ($\mathcal{L}_{\text{MSE}}$) to analyze the impact of difference-aware weighting on the final performance. We report the average classification accuracy with standard deviation and the worst-case classification accuracy under different noise conditions.

\textbf{Analysis.} As shown in Table~\ref{tab:ablation_study}, introducing either $\mathcal{L}_{\text{DCTA}}$ or $\mathcal{L}_{\text{DAC}}$ consistently improves the performance compared with the original baseline model under different noise conditions, demonstrating their effectiveness in enhancing cross-modal representation learning. Moreover, the complete ACML framework achieves the best performance in terms of both average classification accuracy and worst-case classification accuracy, further validating the complementarity between $\mathcal{L}_{\text{DCTA}}$ and $\mathcal{L}_{\text{DAC}}$. Furthermore, replacing $\mathcal{L}_{\text{DCTA}}$ with either $\mathcal{L}_{\text{SC}}$ or $\mathcal{L}_{\text{CSC}}$ leads to inferior performance, demonstrating the superiority of the proposed dynamic cross-modal triplet alignment strategy in regularizing cross-modal representation distributions. Similarly, replacing $\mathcal{L}_{\text{DAC}}$ with the standard unweighted $\mathcal{L}_{\text{MSE}}$ constraint also degrades the performance, validating the effectiveness of the proposed difference-aware weighting mechanism.
\begin{figure}[t!]
\centering
\includegraphics[width=0.65\textwidth]{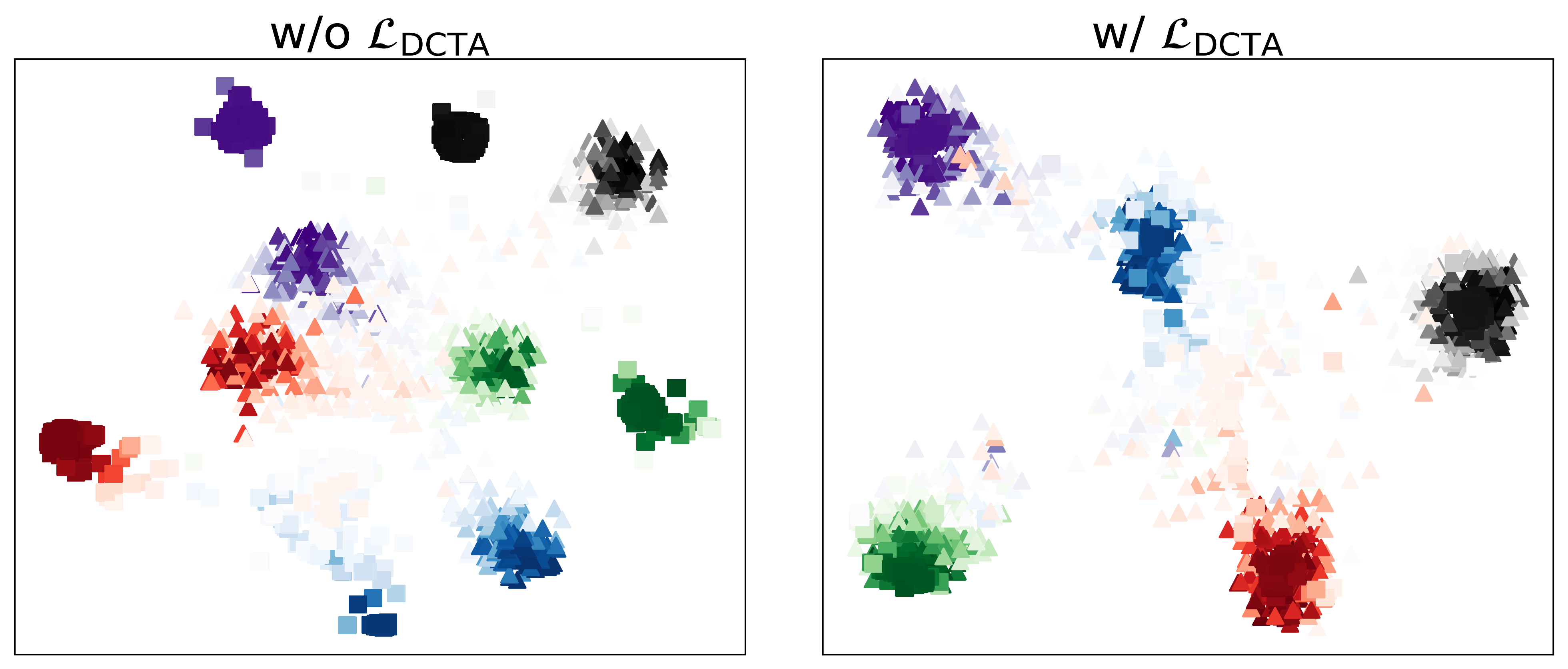}
\caption{2D \(t\)-SNE visualization of multimodal features. \(\triangle\) indicates the image modality, and \(\square\) denotes the text modality. Each color corresponds to a distinct class, with darker shades indicating higher confidence.}
\label{fig:tsne}
\end{figure}
\subsection{RQ3 (Visualization Analysis)}
\textbf{Setting.} To verify whether $\mathcal{L}_{\text{DCTA}}$ effectively reshapes the representation space and improves cross-modal semantic alignment, we perform $t$-SNE visualizations on the FOOD101 dataset to qualitatively examine the feature distributions before and after applying $\mathcal{L}_{\text{DCTA}}$.

\textbf{Analysis.} As shown in Figure~\ref{fig:tsne}, the $t$-SNE visualizations further reveal that, after introducing $\mathcal{L}_{\text{DCTA}}$, high-confidence samples form more compact clusters with improved cross-modal consistency. In contrast, low-confidence samples (indicated by lighter colors) remain relatively scattered, indicating that the proposed loss effectively promotes cross-modal alignment.

\subsection{RQ4 (Hyper-parameter Analysis)}
\textbf{Setting.} We evaluate the sensitivity of two key hyper-parameters, $\lambda_1$ and $\lambda_2$, on the NYU Depth V2 and FOOD101 datasets, which control the constraint strengths of $\mathcal{L}{_\text{DCTA}}$ and $\mathcal{L}{_\text{DAC}}$, respectively. The search spaces for both hyper-parameters are set to $\left \{0.1, 0.5, 1, 5\right \}$. We report the mean classification accuracy under three settings, including the clean setting, $\epsilon=5$, and $\epsilon=10$ noisy settings, where the final result is obtained by averaging the accuracies across the three settings.

\begin{figure}[t!]
\centering
\includegraphics[width=0.75\textwidth]{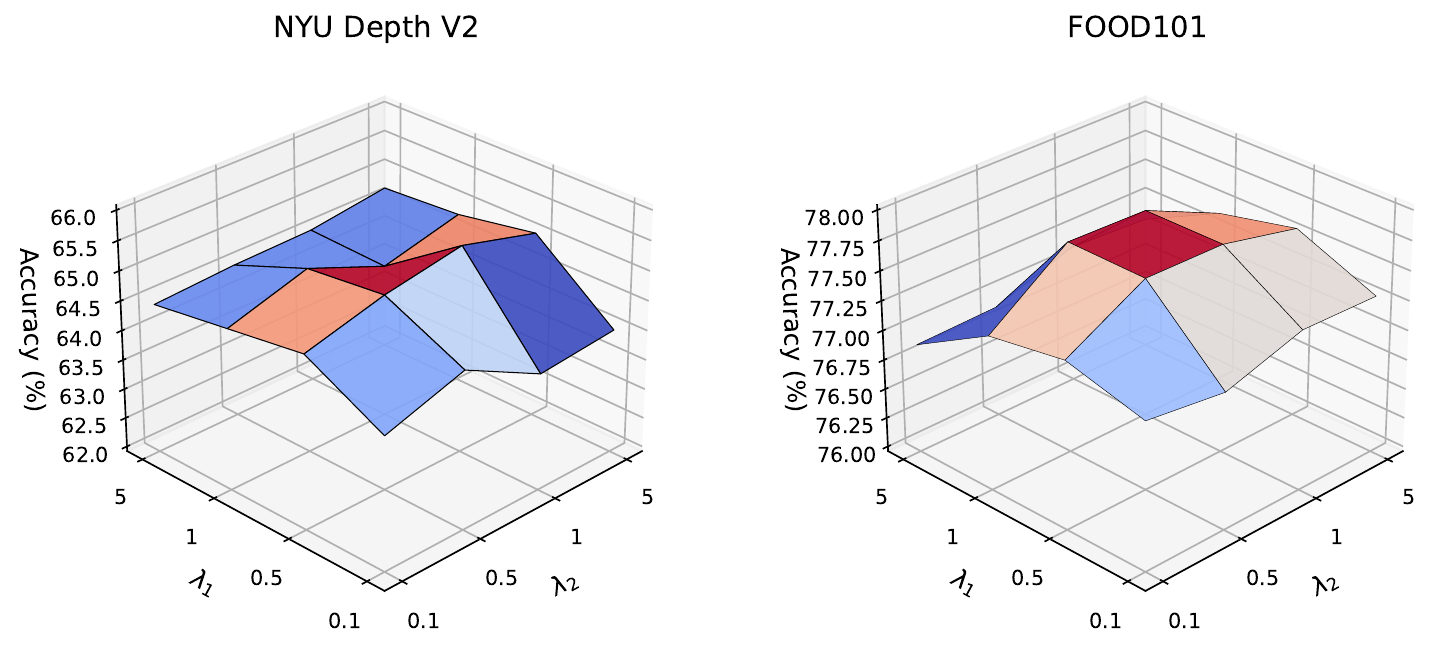}
% \vspace{-20pt}
\caption{Performance sensitivity to hyperparameters $\lambda_1$ and $\lambda_2$ across two datasets.}
\label{fig:hyperparam}
\end{figure}

\textbf{Analysis.} As illustrated in Figure~\ref{fig:hyperparam}, excessively small or large values of either $\lambda_1$ or $\lambda_2$ lead to noticeable performance degradation. In contrast, when the hyper-parameters are within a reasonable range (e.g., $0.5$ and $1$), the proposed method achieves more stable and robust performance. Similar trends can be observed for both hyper-parameters, and the phenomenon becomes more evident on the larger-scale FOOD101 dataset. These results demonstrate that the proposed method maintains good stability and robustness within a reasonable hyper-parameter range.
\begin{table}[t!]
\centering
\caption{Accuracy comparison on FOOD101 validation and test sets under different missing rates (baselines vs. +ACML)}
\small
\resizebox{\textwidth}{!}{
\begin{tabular}{
c|cc|
c>{\columncolor{gray!20}}c|
c>{\columncolor{gray!20}}c|
c>{\columncolor{gray!20}}c|
c>{\columncolor{gray!20}}c
}
\toprule

\multirow{2}{*}{\begin{tabular}[c]{@{}c@{}}Missing\\ rate $\eta$\end{tabular}} 
& \multicolumn{2}{c|}{Train/Test} 
& \multicolumn{4}{c|}{Validation Set} 
& \multicolumn{4}{c}{Testing Set} \\

\cmidrule(lr){2-3}
\cmidrule(lr){4-7}
\cmidrule(lr){8-11}

& Image & Text 
& DCP 
& \scriptsize(+ACML) 
& SyP 
& \scriptsize(+ACML) 
& DCP 
& \scriptsize(+ACML) 
& SyP 
& \scriptsize(+ACML) \\ 

\midrule

\multirow{3}{*}{50\%} 
& 100\% & 50\% 
& 82.33 & \textbf{84.72} 
& 82.64 & \textbf{83.68} 
& 82.11 & \textbf{84.42} 
& 81.93 & \textbf{83.85} \\

& 50\% & 100\% 
& 89.23 & \textbf{89.44} 
& 89.07 & \textbf{89.16} 
& 89.12 & \textbf{89.44} 
& 89.18 & \textbf{89.39} \\

& 75\% & 75\% 
& 85.25 & \textbf{85.86} 
& 85.20 & \textbf{86.28} 
& 85.24 & \textbf{86.03} 
& 86.17 & \textbf{86.50} \\ 

\hline

\multirow{3}{*}{70\%} 
& 100\% & 30\% 
& 79.18 & \textbf{80.61} 
& 79.39 & \textbf{80.21} 
& 78.87 & \textbf{80.39} 
& 79.28 & \textbf{80.27} \\

& 30\% & 100\% 
& 87.53 & \textbf{88.11} 
& 87.62 & \textbf{87.87} 
& 87.32 & \textbf{87.77} 
& 87.27 & \textbf{87.75} \\

& 65\% & 65\% 
& 82.38 & \textbf{83.24} 
& 82.53 & \textbf{83.55} 
& 81.87 & \textbf{83.11} 
& 82.25 & \textbf{83.24} \\ 

\hline

\multirow{3}{*}{90\%} 
& 100\% & 10\% 
& 75.54 & \textbf{77.21} 
& 76.07 & \textbf{77.05} 
& 75.26 & \textbf{77.54} 
& 76.30 & \textbf{77.47} \\

& 10\% & 100\% 
& 86.26 & \textbf{86.36} 
& 86.50 & \textbf{86.59} 
& 85.78 & \textbf{86.00} 
& 86.19 & \textbf{86.55} \\

& 55\% & 55\% 
& 80.39 & \textbf{80.77} 
& 80.64 & \textbf{81.48} 
& 79.87 & \textbf{79.93} 
& 79.31 & \textbf{80.59} \\ 

\bottomrule
\end{tabular}
}
\label{missing}
\end{table}
\subsection{RQ5 (Applicability to Missing Modalities)}
\textbf{Setting.} To evaluate the adaptability of ACML under missing-modality conditions, we integrate it as a plug-and-play module into two prompt-tuning-based approaches, DCP\cite{shi2024deep} and SyP\cite{zhang2025synergistic}. Experiments are conducted on the FOOD101 dataset. Following prior works\cite{shi2024deep,zhang2025synergistic}, we consider three missing ratios: 50\%, 70\%, and 90\%. For each ratio, we further evaluate three missing patterns: text missing, image missing, and both modalities missing, where each modality is removed at a ratio of 50\%. All methods keep the backbone encoders frozen and only update the prompts together with our module. Performance is reported on both the validation and test sets.

% \textbf{Analysis.} As shown in Table \ref{missing}, integrating ACML consistently improves performance under both DCP and SyP frameworks across all missing-ratio settings. This trend holds for 50\%, 70\%, and 90\% missing modalities, indicating that ACML is robust to varying degrees of modality absence.
\textbf{Analysis.} As reported in Table~\ref{missing}, ACML consistently improves the performance of both DCP and SyP across all missing-modality ratios and missing patterns on both validation and test sets. Notably, the gains remain stable even under severe modality absence (e.g., 90\% missing ratio), demonstrating the strong robustness of ACML to incomplete multimodal inputs. These results suggest that ACML can effectively enhance representation learning when modality information is partially unavailable and can be seamlessly integrated into different prompt-tuning frameworks as a general and effective enhancement module.

\subsection{RQ6 (Generalization and Scalability)}
\textbf{Setting.} To evaluate the adaptability of ACML in broader multi-modal scenarios, we conduct classification experiments on two medical multi-omics benchmark datasets: BRCA\cite{wang2021mogonet} and ROSMAP\cite{a2012overview,de2018multi}. These tasks involve three sequential omics modalities ($M=3$) and particularly emphasize population-level relationship modeling. Specifically, the BRCA dataset contains 875 samples spanning five subtypes, while the ROSMAP dataset includes 351 samples for binary Alzheimer’s disease diagnosis. For the network architecture, each modality employs a lightweight encoder—consisting of a single linear layer followed by a ReLU activation—to project input features into a 2048-dimensional space ($d_m=2048$), with an MLP used for feature dimension alignment ($d=2048$). The model is trained for 500 epochs using the Adam optimizer with a learning rate of $10^{-4}$. For evaluation, the BRCA dataset uses accuracy, macro F$_1$-score, and weighted F$_1$-score, while the ROSMAP dataset adopts accuracy, F$_1$-score, and AUC to comprehensively assess diagnostic performance. All results are reported as the mean over 20 random seeds.

\textbf{Analysis.} We compare ACML against six SOTA multimodal methods that focus on sequence-level modeling: \textbf{MOGONet} \cite{wang2021mogonet}, \textbf{MMDyn} \cite{han2022trusted}, \textbf{MLCLNet} \cite{zheng2023multi}, \textbf{DMIB} \cite{fang2024dynamic}, \textbf{GCFANet} \cite{zheng2024global}, and \textbf{HCMAF} \cite{gan2025hcmaf}. As shown in Table \ref{multiomic}, ACML achieves significant and consistent performance gains across all tasks. These results demonstrate that our method maintains strong generalization capability across different data types and more diverse multi-modal scenarios.
\begin{table}[t!]
\small
\centering
\caption{Classification results (\%) on the BRCA and ROSMAP datasets.}
\begin{tabular}{c|ccc|ccc}
\toprule
\multirow{2}{*}{Method} & \multicolumn{3}{c|}{BRCA} & \multicolumn{3}{c}{ROSMAP} \\
\cmidrule(lr){2-4} \cmidrule(lr){5-7}
& Accuracy & Weighted F$_1$ & Macro F$_1$ 
& Accuracy & F$_1$ & AUC \\
\midrule

MOGONet 
& 82.9{\scriptsize±1.8} 
& 82.5{\scriptsize±1.7} 
& 77.4{\scriptsize±1.7} 
& 81.5{\scriptsize±2.3} 
& 82.1{\scriptsize±1.2} 
& 87.4{\scriptsize±1.2} \\

MMDyn 
& 87.7{\scriptsize±0.3} 
& 88.0{\scriptsize±0.5} 
& 84.5{\scriptsize±0.5} 
& 84.2{\scriptsize±1.3} 
& 84.6{\scriptsize±0.7} 
& 91.2{\scriptsize±0.7} \\

MLCLNet 
& 86.4{\scriptsize±1.6} 
& 87.8{\scriptsize±1.7} 
& 82.6{\scriptsize±1.8} 
& 84.4{\scriptsize±1.5} 
& 85.2{\scriptsize±1.5} 
& 89.3{\scriptsize±1.1} \\

DMIB 
& 86.0{\scriptsize±1.6} 
& 86.0{\scriptsize±0.8} 
& 81.6{\scriptsize±0.9} 
& 84.9{\scriptsize±1.8} 
& 85.3{\scriptsize±1.7} 
& 91.6{\scriptsize±0.7} \\

GCFANet 
& 88.6{\scriptsize±1.5} 
& 88.9{\scriptsize±1.6} 
& 85.3{\scriptsize±1.6} 
& 86.3{\scriptsize±1.4} 
& 88.3{\scriptsize±1.6} 
& 91.5{\scriptsize±1.2} \\

HCMAF 
& \underline{89.9{\scriptsize±0.2}} 
& \underline{90.1{\scriptsize±0.2}} 
& \underline{86.4{\scriptsize±0.6}} 
& \underline{88.0{\scriptsize±0.7}} 
& \textbf{87.9{\scriptsize±1.2}} 
& \underline{91.9{\scriptsize±0.7}} \\

\rowcolor{gray!20}
\textbf{ACML} 
& \textbf{90.7{\scriptsize±0.2}} 
& \textbf{91.1{\scriptsize±0.2}} 
& \textbf{88.1{\scriptsize±0.2}} 
& \textbf{88.2{\scriptsize±0.8}} 
& \underline{87.6{\scriptsize±0.8}} 
& \textbf{92.4{\scriptsize±0.7}} \\

\bottomrule
\end{tabular}
\label{multiomic}
\end{table}
%-------------------------------------------------------------------------
\section{Conclusion}
This paper proposes an Alignment- and Calibration-driven Multimodal Learning framework (ACML) to address the limitations of existing methods in static alignment, which overlooks sample-wise differences and fusion strategies lacking difference-aware modeling. Specifically, ACML consists of two key modules: (i) a dynamic cross-modal triplet alignment (DCTA) module, which adaptively adjusts alignment strength according to sample confidence, enforcing cross-modal semantic consistency for high-confidence samples while preserving potential complementary information in low-confidence samples through an orthogonality encouragement strategy; and (ii) a difference-aware attention calibration (DAC) module, which adaptively adjusts the attention regularization strength according to feature magnitude differences and confidence differences across modalities, enabling more reasonable sample-aware fusion while avoiding unnecessary over-regularization. Experimental results demonstrate that ACML consistently achieves superior performance and robustness across multiple multimodal benchmark datasets, validating the effectiveness of the proposed modules. In future work, we plan to explore the integration of ACML with large-scale pre-trained foundation models, extending its applicability to complex real-world scenarios such as medical multimodal analysis, as well as multimodal understanding and generation tasks.

\section*{Acknowledgements}
This work was supported by the National Natural Science Foundation of China (No. 62276052), and the Medico Engineering Cooperation Funds from University of Electronic Science and Technology of China and West China Hospital (No. ZYGX2022YGRH009).
\bibliography{egbib}
\end{document}